# FedXCrop: Privacy-Preserving Crop Disease Detection via Federated Learning with Post-Aggregation Explainability under Non-IID Farm Distributions

[1]Kriti Saini, School of Computer Science and Engineering, Vellore Institute of Technology, Chennai, India – 600127, kriti.saini2023@vitstudent.ac.in

[2] Manikandan P *, School of Computer Science and Engineering, Vellore Institute of Technology, Chennai, India – 600127, manikandan.pr@vit.ac.in.

**ABSTRACT**

Deep learning models identify crop diseases from leaf images with high accuracy on benchmark datasets, but two obstacles stand between these models and field use: training normally requires farms to pool their images on a central server, and a prediction without an explanation gives a farmer or agronomist little basis for a costly treatment decision. This paper presents FedXCrop, a federated learning pipeline for crop disease classification in which explanations of the aggregated global model are evaluated quantitatively rather than by inspection of a few images. A MobileNetV2 classifier trained centrally on PlantVillage reached a test accuracy of 99.38 ± 0.05% (mean ± standard deviation over three seeds) on a fixed test split of 5,428 images covering 38 classes. Federated models were initialized from ImageNet weights and trained for 30 rounds on the full training split distributed across five clients. FedAvg reached 98.81 ± 0.09% with identically distributed clients and 98.81 ± 0.25%, 98.59 ± 0.14% and 97.78 ± 0.48% under Dirichlet label skew with concentration $\alpha$ = 1, 0.5 and 0.1. FedProx, with its proximal coefficient selected on the validation split, was not significantly different from FedAvg at any level of heterogeneity (McNemar $p > 0.3$ in every case). Grad-CAM and SmoothGrad were applied to the aggregated models on 380 test images. Both placed more attribution on the leaf than chance (energy ratio 0.70 to 0.78 against a chance level of 0.60), and Grad-CAM maps of the federated models correlated with those of the centralized model (Spearman $\rho$ 0.74 to 0.76). Agreement fell as heterogeneity increased, attribution on lesion tissue was only modestly above chance, and a model randomization check showed that SmoothGrad maps depend substantially on image structure rather than on the learned classifier. Federated training under label skew therefore cost between 0.6 and 1.6 percentage points of accuracy in this setting and largely preserved Grad-CAM explanations, while proximal regularization gave no measurable benefit.

## 1. Introduction

Plant diseases are estimated to reduce global agricultural yield by 20 to 40 percent each year, which makes them a major threat to food security and sustainable farming. Crop pathogens and pests affect many economically important species, and their impact is most severe for smallholder communities that depend on agricultural production for both income and food. Where smallholder farming dominates, early and accurate diagnosis is limited by access to agronomic expertise, so automated image-based detection has become an important research priority . Convolutional neural networks (CNNs), particularly lightweight transfer-learned architectures, perform strongly on benchmark plant disease datasets, and models trained on PlantVillage routinely exceed 95 percent classification accuracy. Systematic reviews covering more than 160 studies report that transfer-learned architectures such as MobileNetV2, ResNet and EfficientNet achieve state-of-the-art results on leaf image classification.

Despite this benchmark accuracy, two obstacles complicate deployment. First, conventional training requires raw leaf images from geographically distributed farms to be collected on a single server. This raises data ownership and privacy concerns, since farm-level disease data can be commercially sensitive, and limited rural bandwidth can make large transfers impractical. Second, an accurate model is not necessarily an interpretable one: a prediction delivered without visual evidence gives a farmer or agronomist little basis for deciding whether an expensive treatment is justified. The black-box nature of deep learning has been identified as one of the main barriers to the adoption of artificial intelligence in agriculture.

Federated learning (FL) addresses the first obstacle by allowing several farms or agricultural organizations to train a shared model without transferring their images to a central server. Each client trains locally and sends only model updates to an aggregation server. FL has been applied to crop disease identification, yield prediction and irrigation control. Two qualifications apply. Sharing model updates instead of images reduces exposure but does not by itself guarantee privacy, since gradients can leak information about training data. In addition, data held by different farms is rarely identically distributed: farms grow different crops, face different regional disease pressure and capture images under different conditions. The resulting statistical heterogeneity can degrade standard aggregation strategies such as FedAvg. FedProx adds a proximal term to each client's local objective that keeps local updates close to the current global model, with the aim of reducing client drift under heterogeneous data .

Explainable artificial intelligence (XAI) addresses the second obstacle by providing visual evidence for predictions. Grad-CAM and SmoothGrad produce attribution maps that indicate which image regions most influenced a classification. Both have been applied widely to centrally trained plant disease classifiers, and studies report that well-trained models attend to lesion regions when appropriate augmentation is used. Much less is known about how these methods behave after federated aggregation. Averaging parameters across clients with different data distributions may not preserve the attribution patterns of a centrally trained model. A systematic review of 145 FL and XAI studies in precision agriculture published between 2023 and 2025 found that few studies combine accuracy, interpretability and robustness under realistic agricultural conditions, and that quantitative evaluation of post-aggregation explanations is largely missing .

This paper makes four contributions. First, it evaluates FedAvg and FedProx for crop disease classification under a controlled sweep of label heterogeneity (identically distributed clients and Dirichlet concentrations $\alpha$ = 1, 0.5 and 0.1 ), with three seeds per configuration, a single fixed test split, model selection on the validation split only, bootstrap confidence intervals and paired significance tests. Second, it applies Grad-CAM and SmoothGrad to the aggregated global models and evaluates the resulting explanations quantitatively: faithfulness by deletion and insertion curves , localization against leaf masks and lesion pseudo-masks, agreement with the explanations of a centrally trained model, and a model randomization sanity check. Third, it reports where these explanations fall short, including attribution failure cases, the limited lesion-level localization and the partial failure of SmoothGrad under randomization. Fourth, it shows that a common shortcut in federated plant disease experiments, initializing clients from a model already trained centrally on the same data, yields accuracies that do not reflect federated training. All results are produced by publicly released code and are traceable to stored result files.

## 2. Related Work

Deep learning, and CNNs in particular, has been used widely for plant disease detection. Mohanty et al. [2] were among the first to show that deep CNNs trained on PlantVillage could exceed 99 percent accuracy, demonstrating that leaf images carry enough visual information for accurate diagnosis [8]. Later studies applied transfer learning with ResNet, VGG, DenseNet and several MobileNet variants and reported similarly strong results [16,48]. Comparisons between deep learning and traditional machine learning found that deep learning performs better, particularly as the number of disease classes and training images grows [10].

MobileNet architectures have received particular attention because they balance accuracy against computational cost. A review of 160 studies published between 2020 and 2024 found MobileNet models among the most popular choices for plant disease detection, owing to their suitability for devices with limited computing resources [17]. Transfer learning from ImageNet weights has also produced good results on relatively small plant disease datasets [48]. Hybrid designs have been explored as well: combining MobileNetV2 with residual blocks reached 99.45 percent accuracy on PlantVillage while remaining light enough for mobile use [18], and Vision Transformers combined with explainability techniques have performed competitively, particularly for multi-label disease classification [19].

Strong results on laboratory datasets do not guarantee equivalent performance in the field. Reviews identify limited dataset diversity, the gap between controlled and field images, and the lack of interpretability as persistent challenges [9,14,20,42], and these need to be addressed before such systems can be relied on in practice.

### 2.1 Federated Learning in Agriculture

Centralized training requires images from participating farms to be transferred to a common server. In agriculture this is difficult because farm data may be sensitive, distributed across many sites, or expensive to transfer over limited connectivity. Federated learning lets each client train locally and send only model updates to the server [3,38].

Kabala et al. [7] evaluated CNN and Vision Transformer models with three federated strategies (FedAvg, FedProx and SMC-Avg) on PlantVillage and found that federated models approached centralized performance while keeping data on the clients. Hari et al. [25] proposed an improved federated deep learning approach for plant leaf disease detection. Other studies have examined FL in smart farming [21,47], rice disease detection on resource-constrained IoT devices [22], wheat disease severity assessment [46] and soybean disease detection [40,50]. These studies cover different crops and deployment settings but do not jointly examine client heterogeneity, predictive performance and explanation quality. More specialized approaches include knowledge sharing between regional models without pooling training data [11] and federated object detection for multiple crop diseases and pests [35], which shows that agricultural FL is moving beyond single-disease classification.

Federated learning also introduces its own challenges. Farms can hold very different types and amounts of data, producing non-IID distributions across clients, and communication between clients and server can be costly where connectivity is poor. Reviews of FL in smart agriculture identify privacy, communication efficiency and robustness to non-IID data as recurring deployment concerns [39,49], and the review of 145 FL and XAI studies cited above found that few studies address accuracy, interpretability and robustness to heterogeneity together [23]. Studies of heterogeneous farm environments have shown that Dirichlet-based partitioning represents differences between farms more faithfully than random splits, and that FedProx can be more stable than FedAvg when heterogeneity is high [24]. Dirichlet label partitioning was introduced as a controlled way to vary heterogeneity in federated image classification [51].

### 2.2 Explainable AI for Plant Disease Models

As deep models have grown more complex, attention has turned to understanding why they make particular predictions. In agriculture this matters because farmers and experts need to know which regions of a leaf influenced a diagnosis, and whether the model is responding to visible symptoms or to unrelated image features.

Grad-CAM [5] is one of the most widely used methods for visual explanations of CNNs and has been applied to crop disease datasets involving tomato, potato, apple and corn [26,41]. A review of 427 studies published between 2020 and 2024 reported its widespread use in agricultural and medical imaging [27]. The same review notes that Grad-CAM maps are not always reliable indicators of a model's decision process: a model may attend to background regions instead of symptoms, particularly when training data lacks variation [27]. SHAP [12] and SmoothGrad [6] offer further ways to examine the features that drive a prediction, and combining methods such as Grad-CAM and SHAP has been reported to give more dependable interpretations than any single method [41].

Explanations can help experts recognize when a model attends to an irrelevant part of an image, which assists in investigating incorrect predictions [28], and they can indicate whether a decision is associated with disease-related features [19,26,42]. This is especially relevant when systems move from laboratory datasets to field conditions, where backgrounds, lighting and leaf orientation differ from the training data [43]. Visual plausibility is not the same as faithfulness, however. Adebayo et al. [55] showed that some attribution methods produce nearly identical maps for trained and randomly initialized networks, so their output can reflect image structure rather than what the model learned. Quantitative checks such as deletion and insertion curves [53], the pointing game [54] and randomization tests [55] are therefore needed to support claims about explanation quality. Most existing XAI studies for plant disease detection use centrally trained models and qualitative inspection [23,30].

### 2.3 Explainability in Federated Learning

Combining XAI with federated learning is a young research area. Although FL keeps data distributed, the same decentralized structure makes it harder to understand how individual local updates contribute to the global model [34]. Initial work in medical imaging applied Grad-CAM to a federated brain tumor classifier after aggregation and found that FedProx, particularly when combined with data augmentation, preserved spatial explanation quality better than FedAvg under non-IID conditions [29]. Reviews of XAI and FL in medical imaging conclude that integrating explainability into federated systems is at an early stage and that standardized evaluation is not yet established [15,30].

Non-IID data is a central consideration. When client datasets differ substantially, FedAvg can converge slowly or inconsistently [31,45]. FedProx limits how far a client model can move from the current global model during local training [4,13,32], which can stabilize training when client distributions differ. Privacy and explainability can also be in tension: some explanation techniques require access to model outputs or intermediate representations that could expose information about the data or the model [44].

### 2.4 Research Gap

Considerable progress has been made in both federated crop disease detection [7,21,22,24,46,50] and explainable plant disease classification [26,27,28,41,42], but the two lines of work have developed largely independently. Federated agriculture studies concentrate on classification accuracy and rarely analyze how the resulting models reach their predictions [21,22,46,50], while XAI studies in plant disease detection use centrally trained models [26,27,28,41]. Kabala et al. [7] compared FedAvg and FedProx on PlantVillage without explanations, and Albalawi et al. [29] studied post-aggregation Grad-CAM in medical imaging. A 2026 systematic review identified the lack of studies that apply and quantitatively evaluate post-aggregation XAI in federated crop disease detection under non-IID conditions [23]. FedXCrop addresses this gap by evaluating the explanations of aggregated models with faithfulness, localization, agreement and sanity metrics across controlled levels of heterogeneity, alongside a statistically tested comparison of classification performance.

## 3. Materials and Methods

Figure 1 summarizes the pipeline. Every experiment uses one fixed data split, every federated model starts from ImageNet weights, and every configuration is scored on the same held-out test images.

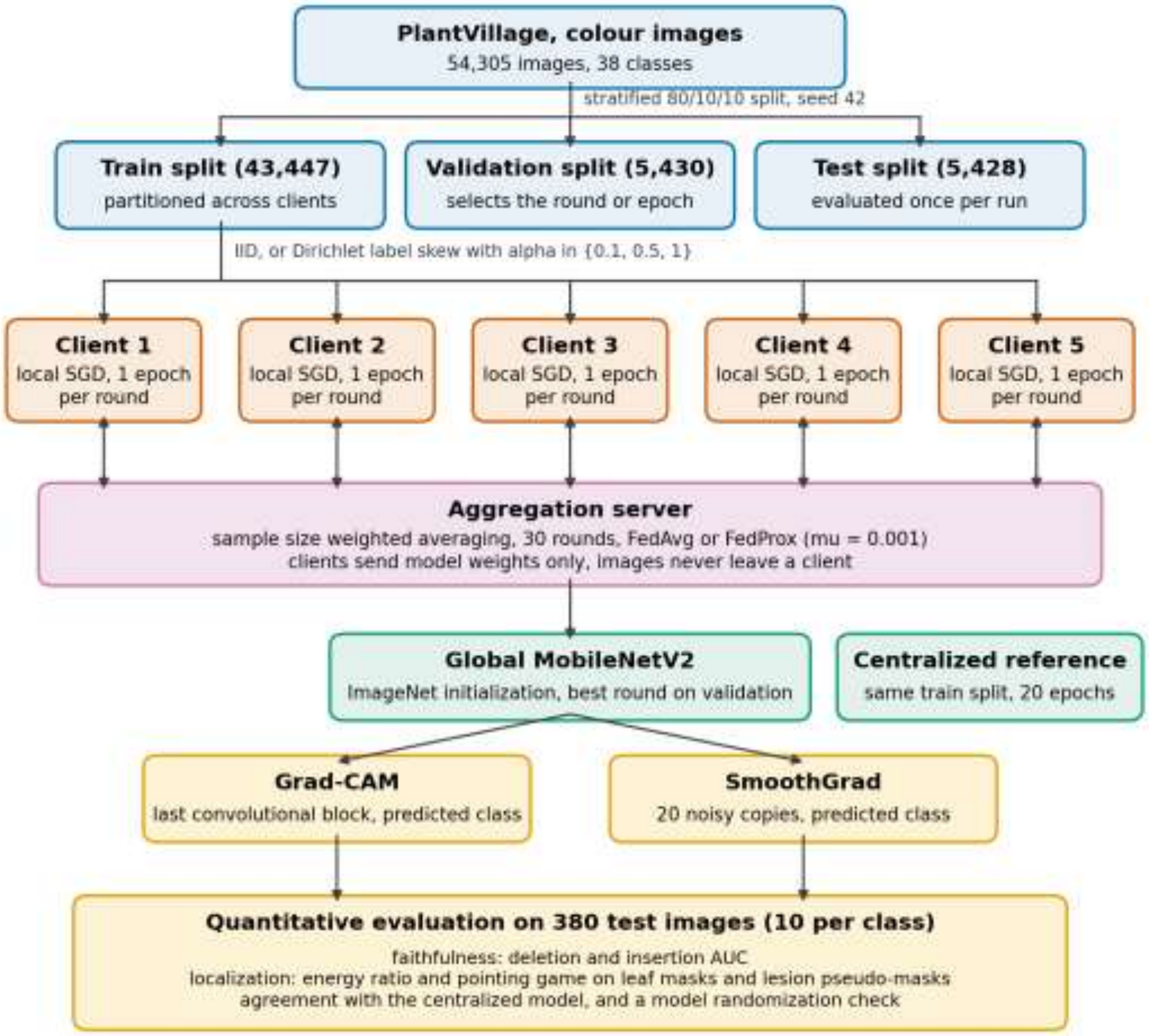

**Figure 1.** Overview of FedXCrop. The training split is partitioned across five clients, which train locally and send only model weights to the aggregation server. The validation split selects the round (or epoch) carried forward, the test split is evaluated once per run, and explanations of the global model are evaluated quantitatively on a fixed sample of test images. All numbers in the diagram are read from the stored run configuration and result files.

### 3.1 Dataset and Split

Experiments use the colour variant of PlantVillage [2,61], which contains 54,305 leaf images in 38 classes (26 diseases and 12 healthy classes) across 14 plant species, including tomato, potato, apple, corn and grape. Class sizes are imbalanced, from 152 images (potato, healthy) to 5,507 images (orange, Huanglongbing). The data was divided once by stratified random sampling with seed 42 into training (43,447 images), validation (5,430) and test (5,428) splits, so that class proportions are preserved in each split. The split files are stored with the code, and no later step resamples them.

Because PlantVillage contains several photographs of the same leaves, near-duplicate images can appear on both sides of a random split. A perceptual hash comparison (64-bit pHash, Hamming distance at most 4) found that 157 of the 5,428 test images (2.89 percent) have a near-duplicate in the training split, 122 of them with the same label. The split was not modified, so that results remain comparable with published PlantVillage numbers that are subject to the same overlap, but this overlap should be kept in mind when reading absolute accuracies.

For evaluation, images were resized to 256 pixels on the shorter side, center-cropped to 224 × 224 and normalized with ImageNet statistics. Training images were augmented with RandomResizedCrop (224, scale 0.6 to 1.0), random horizontal and vertical flips, random rotation of up to 45 degrees and ColorJitter (brightness 0.4, contrast 0.4, saturation 0.3, hue 0.1). PlantVillage images share a consistent leaf position and a uniform background, which a model can exploit as a shortcut [27,28]; the spatial augmentations are intended to discourage this, and augmentations of this kind are standard practice in agricultural deep learning [17,36]. Augmentation was applied only to the training split.

### 3.2 Model and Centralized Baseline

The classifier is MobileNetV2 [1] initialized with ImageNet weights, with the final layer replaced by a linear layer mapping the 1,280-dimensional feature vector to 38 classes. All layers were trained, so that the backbone could adapt from ImageNet features to leaf imagery [18,48]. The centralized baseline was trained on the full training split for 20 epochs with the Adam optimizer [37] (learning rate 0.001, batch size 32) and a step schedule that multiplies the learning rate by 0.1 every 7 epochs. The epoch with the highest validation accuracy was retained (the 16th, 18th and 20th epochs for seeds 0, 1 and 2), and the test split was evaluated once with that model. Training curves are shown in Supplementary Figure S1.

### 3.3 Federated Learning

Federated training was simulated with Flower [33] using K = 5 clients, all of which participate in every round. The full training split is partitioned across the clients, so the federated and centralized models are trained on the same images. Validation and test splits stay global, and every configuration is scored on identical images.

**Partitions.** In the IID configuration the training images are assigned to clients uniformly at random. In the non-IID configurations each class is divided among clients according to proportions drawn from a symmetric Dirichlet distribution with concentration $\alpha$ [51]; smaller $\alpha$ produces stronger label skew. Three concentrations were used, $\alpha \in \{0.1, 0.5, 1\}$, each with three partition seeds. Heterogeneity was measured by the Jensen-Shannon divergence [52] between each client's label distribution and that of the full training split. Table 1 and Figure 2 summarize the seed 0 partitions. At $\alpha$ = 0.1, clients hold between 4,928 and 15,156 images, and the most skewed client sees only 19 of the 38 classes.

*Table 1. Client partitions of the 43,447-image training split (seed 0). JSD is the Jensen-Shannon divergence between a client's label distribution and the global one.*

| Partition | Client size (min to max) | Classes per client (min to max, mean) | Mean JSD (min to max) |
|---|---|---|---|
| IID | 8,689 to 8,690 | 38 to 38, 38.0 | 0.001 (0.001 to 0.001) |
| Dirichlet $\alpha$ = 1 | 6,815 to 10,775 | 37 to 38, 37.6 | 0.120 (0.099 to 0.146) |
| Dirichlet $\alpha$ = 0.5 | 5,859 to 12,205 | 32 to 38, 35.8 | 0.254 (0.173 to 0.374) |
| Dirichlet $\alpha$ = 0.1 | 4,928 to 15,156 | 19 to 38, 25.4 | 0.470 (0.277 to 0.596) |

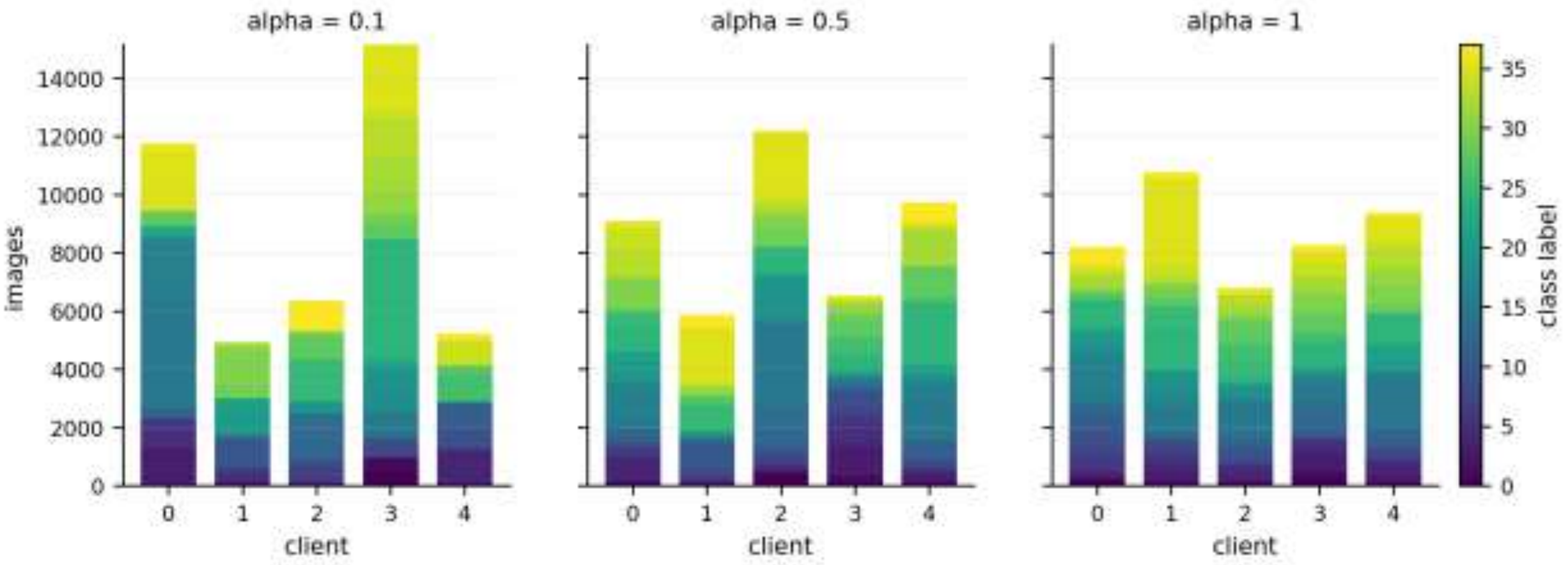

**Figure 2.** Class composition of the five clients under Dirichlet partitions with $\alpha = 0.1$, 0.5 and 1 (seed 0). Each bar is one client, stacked by class label. Lower $\alpha$ concentrates each class on fewer clients and makes client sizes more unequal.

**Training.** Every client and the global model start from the same ImageNet-initialized MobileNetV2. In each round, each client trains for one local epoch with SGD (learning rate 0.01, momentum 0.9, batch size 32), with the optimizer re-created at the start of the round, and returns its weights. The server averages client weights in proportion to client sample counts (FedAvg [3]). FedProx [4] uses the same aggregation but adds a proximal term to each client's objective, $h_k(w) = F_k(w) + (\mu/2)\|w - w^t\|^2$, where $F_k$ is the local loss and $w^t$ the current global model. Training runs for 30 rounds. After every round the global model is evaluated on the full validation split, the round with the highest validation accuracy is retained, and the test split is evaluated once with that model. Each configuration was run with seeds 0, 1 and 2.

**Selection of μ.** The proximal coefficient was chosen from {0.001, 0.01, 0.1} by validation accuracy at $\alpha = 0.1$ (seed 0), without consulting the test split. Validation accuracy was 98.03 percent for $\mu = 0.001$, 97.73 percent for $\mu = 0.01$ and 97.07 percent for $\mu = 0.1$, so $\mu = 0.001$ was used for all FedProx runs.

**Communication.** MobileNetV2 with a 38-class head has 2,272,550 parameters. At 32-bit precision, one round with five clients exchanges about 90.9 MB (download and upload), or about 2.73 GB over 30 rounds.

**Centralized initialization (preliminary protocol).** In preliminary experiments, clients were initialized from the centralized model rather than from ImageNet weights, and only a 5,000-image subset of the training split was distributed to clients for five rounds (Dirichlet $\alpha = 0.5$). Because that initial model has already been trained on every image the clients hold, this protocol does not measure federated learning. It is reproduced here (seed 0) to quantify its effect.

### 3.4 Post-Aggregation Explanations

Explanations are computed on the aggregated global model, which is the model a deployment would use, rather than on individual client models. Both methods were implemented with Captum [58] and explain the predicted class, since the question is what the model used to reach the answer it gave.

**Grad-CAM** [5] was computed at the final convolutional block of MobileNetV2 (model.features[-1]), whose 7 × 7 feature map is the last spatial representation before global average pooling. Gradients of the class score are averaged over spatial positions to weight the activation channels, a ReLU keeps only positive evidence, and the map is upsampled bilinearly to 224 × 224 and normalized to [0, 1]. **SmoothGrad** [6] averages input gradients over 20 noisy copies of each image. The noise standard deviation is 0.1 times the range of the normalized input tensor of that image (about 0.44 in normalized units). Absolute gradients are averaged over colour channels and normalized to [0, 1].

### 3.5 Quantitative Evaluation of Explanations

A fixed stratified sample of 10 test images per class (380 images) was drawn once with seed 42 and used for every model. Explanations were evaluated for the centralized model and for FedAvg and FedProx at $\alpha = 0.1$ and $\alpha = 0.5$ (seed 0 runs).

**Faithfulness.** Deletion and insertion curves [53] rank pixels by attribution and change them in steps of 2 percent of the image. Deletion replaces the most important pixels first with the dataset channel mean; insertion starts from a heavily blurred image (Gaussian $\sigma = 11$ pixels) and restores the most important pixels first. The area under the curve of the predicted class probability is recorded. A faithful map gives a low deletion AUC and a high insertion AUC.

**Leaf localization.** PlantVillage includes a segmented copy of each photograph with the background removed, from which a leaf mask is derived and aligned with the model input through the same resize and crop. Two measures are reported: the energy ratio (the share of total attribution inside the mask) and the pointing game [54] (whether the highest-attribution pixel falls inside the mask). A map that highlights nothing in particular would score the mask's area fraction on both, so the mean leaf area fraction of 0.601 is the chance level.

**Lesion localization.** No lesion annotations exist for PlantVillage, so lesion pseudo-masks were derived by colour. The hue range of healthy tissue was measured from leaf pixels of 200 images of the healthy classes (central 90 percent, hue 39 to 83 on the OpenCV scale of 0 to 179). Leaf pixels outside this range on diseased leaves are labelled as lesion, after eroding the leaf mask to remove blended boundary pixels and applying a morphological opening to remove speckle. Pseudo-masks were computed for the 26 disease classes; 250 of the 260 sampled diseased images had a non-empty mask, with a mean lesion area fraction of 0.156 (the chance level). These masks are a heuristic, not ground truth. Figure 3 shows them for a random sample of 30 diseased images, so their quality can be judged directly, and their limitations are discussed further in Section 5.

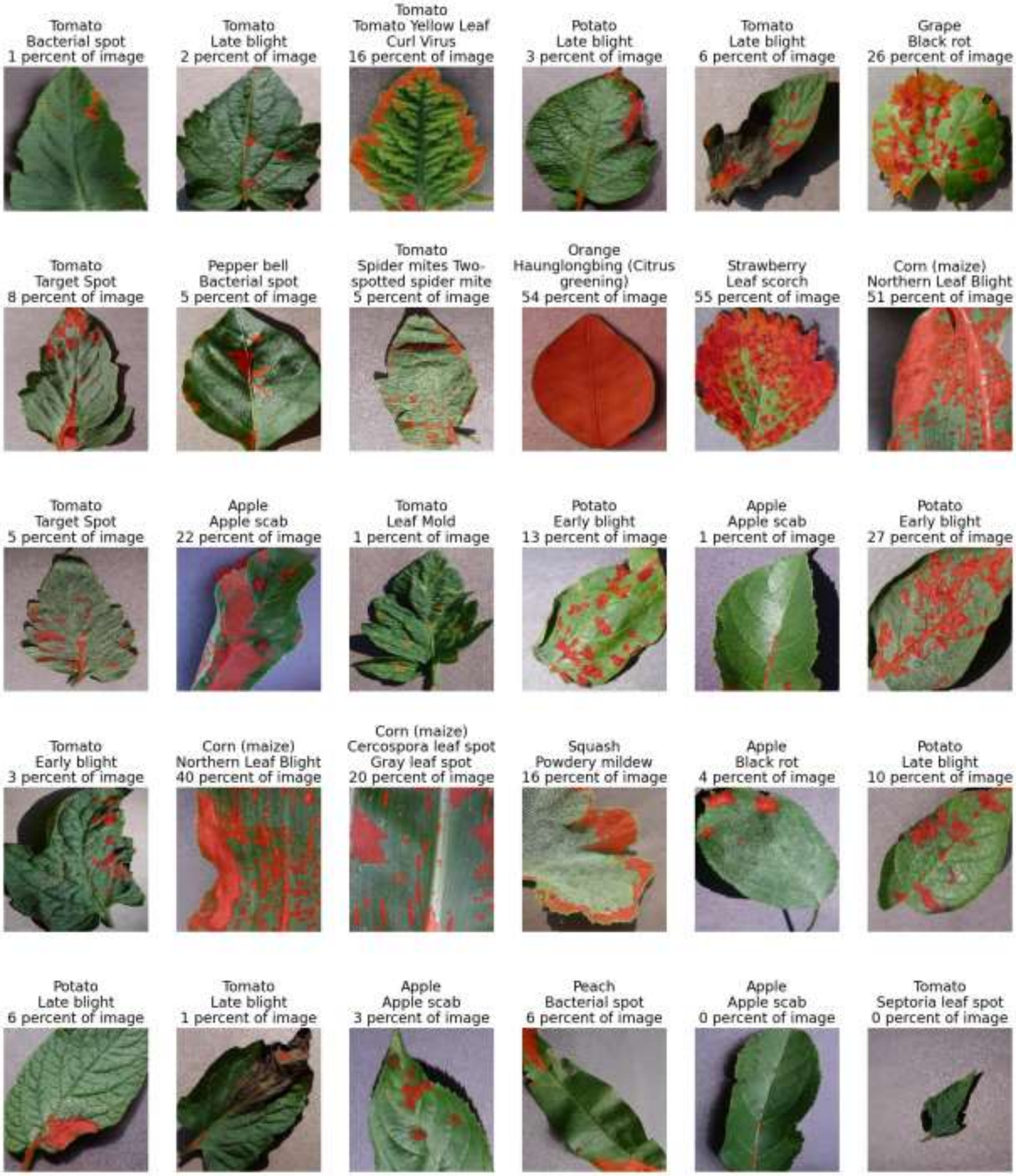

**Figure 3.** Lesion pseudo-masks (red) for 30 randomly chosen diseased images from the explanation sample, with the share of the image each mask covers. The masks follow discrete spots on many leaves, for example grape black rot, potato early blight and apple black rot. They fail in three recurring ways: they mark whole leaves whose overall colour changes (orange Huanglongbing, 54 percent of the image), they miss dark necrotic tissue whose hue falls within the healthy range (tomato late blight), and they are nearly empty for some leaves with small or faint lesions (one apple scab and one tomato Septoria leaf spot image round to 0 percent).

**Agreement with the centralized model.** For each federated model, the Spearman rank correlation and the intersection over union of the top 20 percent of pixels are computed between its map and the centralized model's map for the same image.

**Sanity check.** Following Adebayo et al. [55], the centralized model's weights were randomized cumulatively from the classifier back through the feature blocks, and the Spearman correlation between the original map and the map of the progressively randomized model was recorded. A method whose maps stay similar after randomization is describing the image rather than the model. This check used the first 40 images of the sample, which belong to the four apple classes.

### 3.6 Statistical Analysis

Classification performance is reported as accuracy and macro-averaged precision, recall and F1 over the 38 classes, as mean ± standard deviation across the three seeds. For the seed 0 run of each configuration, a 95 percent bootstrap confidence interval over test images (1,000 resamples) [63] is also reported, along with the number of correctly classified images. Pairs of configurations were compared with McNemar's test [56] (with continuity correction) on the seed 0 predictions, pairing runs from the same seed so that both use the same client partition. Explanation metrics are reported as means over the sampled images with 95 percent bootstrap intervals, and FedProx was compared with FedAvg using paired Wilcoxon signed-rank tests [57] on identical images. The 32 explanation tests were not corrected for multiple comparisons, so individual p-values near 0.05 should be interpreted with caution.

## 4. Results

### 4.1 Experimental Setup

The pipeline was implemented in PyTorch 2.4.1 [59] and torchvision 0.19.1, with Flower 1.11.1 for federated simulation, Captum 0.7.0 for attributions and scikit-learn 1.5.2 for metrics. Experiments ran on Google Colab with a single CUDA GPU. Colour augmentation was applied to whole batches on the GPU rather than in the data-loading workers, using the same per-image random factors as torchvision; on a two-core runtime this raised input throughput from 106 to 272 images per second (a factor of 2.55). Each run's metric files record the number of correct predictions and the number of test images, so every reported rate can be checked against whole counts.

### 4.2 Classification Performance

Table 2 reports test performance for all configurations. The centralized model reached 99.38 ± 0.05 percent accuracy and 99.00 ± 0.07 percent macro F1. Its weakest classes were potato, healthy (recall 86.67 percent, 13 of only 15 test images), corn Cercospora leaf spot (recall 88.46 percent, with all six errors predicted as corn Northern Leaf Blight) and tomato early blight (recall 93.00 percent).

*Table 2. Test performance on the 5,428-image test split. Values are mean ± standard deviation over seeds 0, 1 and 2 (percent). The last two columns give the seed 0 accuracy with its 95% bootstrap interval and the number of correct predictions. FedProx uses $\mu$ = 0.001 unless stated; rows with one seed come from the $\mu$ selection sweep or the preliminary protocol.*

| Configuration | Seeds | Accuracy | Macro F1 | Macro precision | Macro recall | Seed 0 accuracy [95% CI] | Correct (seed 0) |
|---|---|---|---|---|---|---|---|
| Centralized | 3 | 99.38 ± 0.05 | 99.00 ± 0.07 | 99.32 ± 0.03 | 98.73 ± 0.09 | 99.34 [99.10, 99.52] | 5,392 |
| FedAvg, IID | 3 | 98.81 ± 0.09 | 98.37 ± 0.15 | 98.62 ± 0.15 | 98.20 ± 0.18 | 98.89 [98.62, 99.17] | 5,368 |
| FedAvg, $\alpha$ = 1 | 3 | 98.81 ± 0.25 | 98.35 ± 0.37 | 98.66 ± 0.44 | 98.11 ± 0.31 | 98.93 [98.64, 99.19] | 5,370 |
| FedAvg, $\alpha$ = 0.5 | 3 | 98.59 ± 0.14 | 97.99 ± 0.21 | 98.28 ± 0.12 | 97.80 ± 0.27 | 98.64 [98.31, 98.93] | 5,354 |

| Configuration | Seeds | Accuracy | Macro F1 | Macro precision | Macro recall | Seed 0 accuracy [95% CI] | Correct (seed 0) |
|---|---|---|---|---|---|---|---|
| FedAvg, $\alpha = 0.1$ | 3 | 97.78 ± 0.48 | 97.09 ± 0.55 | 97.63 ± 0.32 | 96.75 ± 0.68 | 98.08 [97.72, 98.47] | 5,324 |
| FedProx, $\alpha = 1$ | 3 | 98.73 ± 0.20 | 98.26 ± 0.31 | 98.63 ± 0.15 | 98.00 ± 0.40 | 98.82 [98.53, 99.08] | 5,364 |
| FedProx, $\alpha = 0.5$ | 3 | 98.69 ± 0.10 | 98.15 ± 0.11 | 98.33 ± 0.16 | 98.04 ± 0.11 | 98.66 [98.34, 98.93] | 5,355 |
| FedProx, $\alpha = 0.1$ | 3 | 97.88 ± 0.27 | 97.19 ± 0.32 | 97.73 ± 0.11 | 96.86 ± 0.45 | 98.14 [97.79, 98.49] | 5,327 |
| FedProx, $\alpha = 0.1$, $\mu = 0.01$ | 1 | 97.92 | 97.31 | 97.80 | 97.02 | 97.92 [97.57, 98.29] | 5,315 |
| FedProx, $\alpha = 0.1$, $\mu = 0.1$ | 1 | 97.40 | 96.40 | 97.34 | 95.86 | 97.40 [96.96, 97.79] | 5,287 |
| FedAvg, $\alpha = 0.5$, centralized init. (preliminary protocol) | 1 | 98.91 | 98.34 | 98.97 | 97.90 | 98.91 [98.64, 99.17] | 5,369 |

**Cost of federation.** With identically distributed clients, FedAvg reached 98.81 ± 0.09 percent, 0.57 percentage points below the centralized model. The difference is significant on the seed 0 predictions (the IID model alone was correct on 8 images and the centralized model alone on 32; McNemar $p = 0.0003$). Federated training therefore does not match centralized training here, even without heterogeneity. Table 3 lists all paired comparisons.

*Table 3. Paired McNemar tests on the seed 0 test predictions (5,428 images). Both models in each row use the same seed and therefore the same client partition. "Only A" and "Only B" count the images that only model A or only model B classified correctly; the remaining images carry no information about which model is better. Tests use the chi-square statistic with continuity correction.*

| Model A | Model B | Acc. A (%) | Acc. B (%) | Only A | Only B | p | p < 0.05 |
|---|---|---|---|---|---|---|---|
| FedAvg, IID | Centralized | 98.89 | 99.34 | 8 | 32 | 0.0003 | yes |
| FedAvg, $\alpha = 1$ | FedAvg, IID | 98.93 | 98.89 | 21 | 19 | 0.8744 | no |
| FedAvg, $\alpha = 0.5$ | FedAvg, IID | 98.64 | 98.89 | 28 | 42 | 0.1202 | no |
| FedAvg, $\alpha = 0.1$ | FedAvg, IID | 98.08 | 98.89 | 28 | 72 | < 0.0001 | yes |
| FedProx, $\alpha = 1$, $\mu = 0.001$ | FedAvg, $\alpha = 1$ | 98.82 | 98.93 | 13 | 19 | 0.3768 | no |
| FedProx, $\alpha = 0.5$, $\mu = 0.001$ | FedAvg, $\alpha = 0.5$ | 98.66 | 98.64 | 18 | 17 | 1.0000 | no |
| FedProx, $\alpha = 0.1$, $\mu = 0.001$ | FedAvg, $\alpha = 0.1$ | 98.14 | 98.08 | 38 | 35 | 0.8149 | no |
| FedProx, $\alpha = 0.1$, $\mu = 0.01$ | FedAvg, $\alpha = 0.1$ | 97.92 | 98.08 | 37 | 46 | 0.3799 | no |
| FedProx, $\alpha = 0.1$, $\mu = 0.1$ | FedAvg, $\alpha = 0.1$ | 97.40 | 98.08 | 33 | 70 | 0.0004 | yes |

**Effect of heterogeneity.** Accuracy decreased as label skew increased, from 98.81 ± 0.25 percent at $\alpha = 1$ to 98.59 ± 0.14 percent at $\alpha = 0.5$ and 97.78 ± 0.48 percent at $\alpha = 0.1$ (Figure 4). At $\alpha = 0.1$ accuracy was 1.03 points below IID FedAvg and 1.60 points below the centralized model, and macro recall fell more than macro precision (96.75 against 97.63 percent), meaning that some classes were missed more often rather than over-predicted. Only the strongest skew produced a detectable difference from IID training on the seed 0 predictions (at $\alpha = 0.1$ the skewed model alone was correct on 28 images and the IID model alone on 72, $p < 0.0001$; $\alpha = 0.5$: $p = 0.12$; $\alpha = 1$: $p = 0.87$). Variation across seeds was also largest at $\alpha = 0.1$, where the three partitions gave test accuracies of 98.08, 98.03 and 97.22 percent for FedAvg. Figure 5 shows that heterogeneity mainly slowed early convergence: after the first round, mean validation accuracy was 44.4 percent (FedAvg) and 49.5 percent (FedProx) at $\alpha = 0.1$, compared with 72.7 to 83.1 percent at $\alpha = 0.5$ and $\alpha = 1$. In all settings the federated curves level off below the centralized model's best validation accuracy.

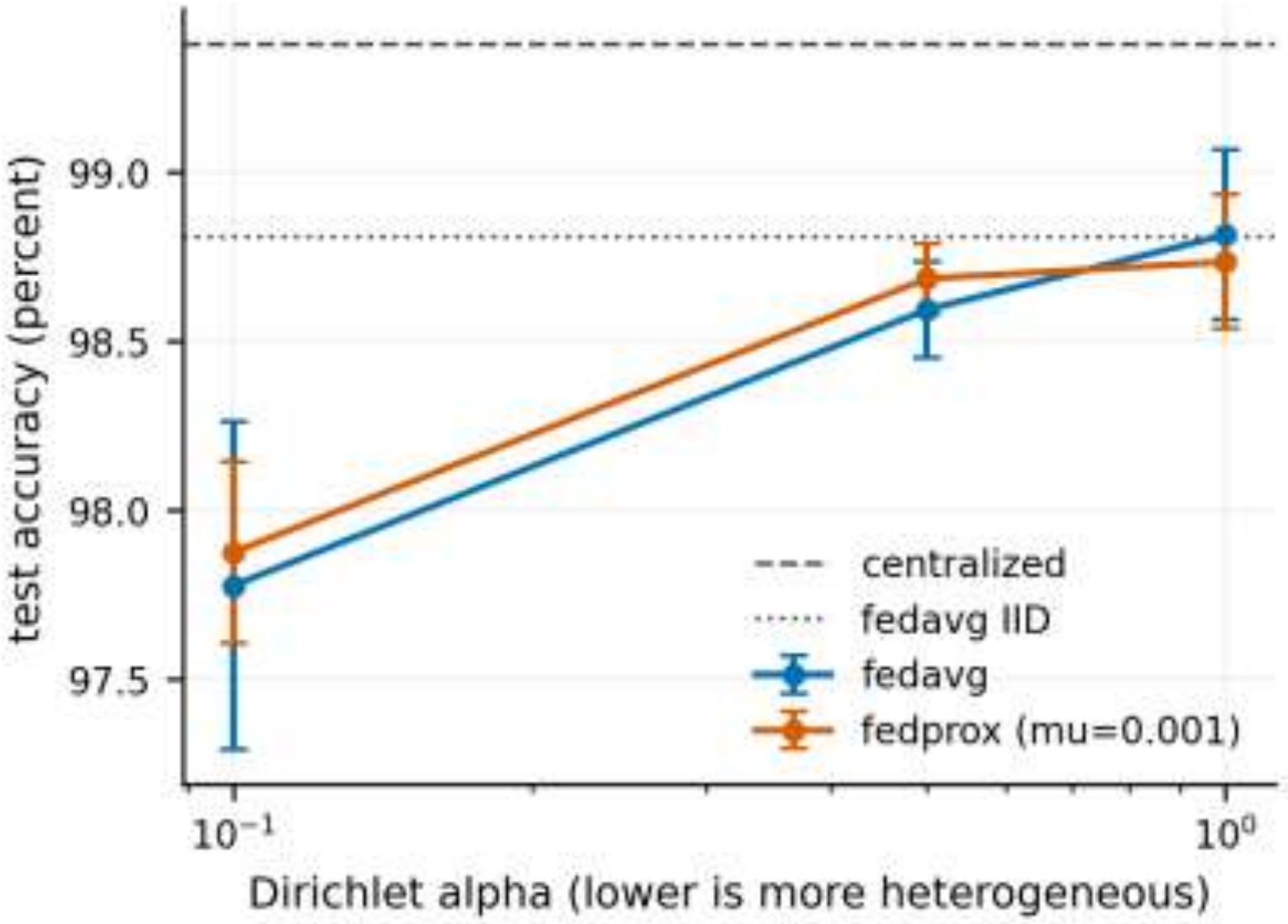

**Figure 4.** Test accuracy against Dirichlet concentration $\alpha$ (log scale; lower $\alpha$ is more heterogeneous). Points are means over three seeds and bars one standard deviation. Horizontal lines mark the centralized model and IID FedAvg.

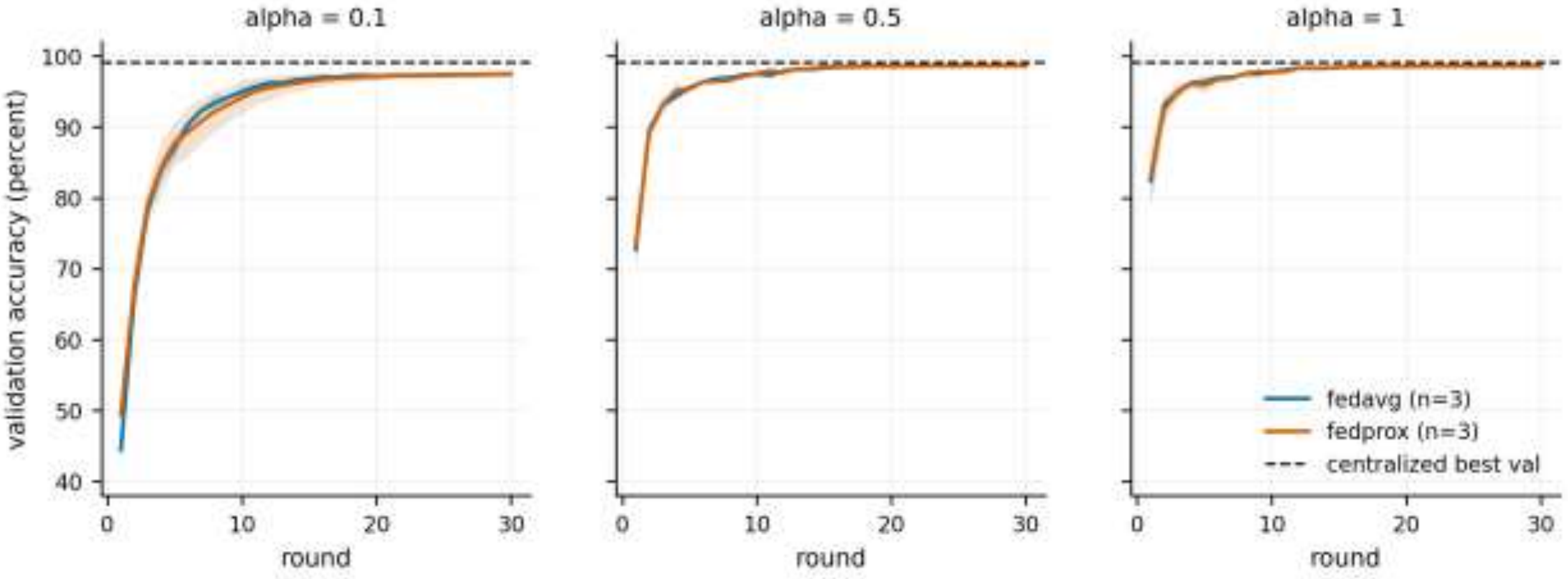

**Figure 5.** Validation accuracy of the global model after each round for FedAvg and FedProx ($\mu = 0.001$) at three levels of label skew. Lines are means over three seeds and bands one standard deviation. The dashed line is the centralized model's best validation accuracy (mean over three seeds).

**FedProx against FedAvg.** FedProx with $\mu = 0.001$ changed mean accuracy by +0.10, +0.09 and −0.08 percentage points at $\alpha = 0.1$, 0.5 and 1. These differences are smaller than the seed-to-seed variation, and none is significant on the seed 0 predictions (McNemar $p = 0.81$, 1.00 and 0.38). At $\alpha = 0.1$, a larger coefficient did not help: $\mu = 0.01$ gave 97.92 percent ($p = 0.38$ against FedAvg on the same partition) and $\mu = 0.1$ gave 97.40 percent, which is significantly worse than FedAvg on the same partition (98.08 percent; FedProx alone was correct on 33 images and FedAvg alone on 70, $p = 0.0004$). The validation curves of the two methods overlap at every level of heterogeneity (Figure 5). With one local epoch per round and full client participation, client drift appears too limited in this setting for the proximal term to help, and larger coefficients mainly slow local progress.

**Preliminary protocol.** When clients were initialized from the centralized model and trained on a 5,000-image subset for five rounds, the global model reached 98.91 percent (5,369 of 5,428), higher than the mean accuracy of every ImageNet-initialized federated configuration. This accuracy is inherited from centralized training on the full training split, a model that could not exist in the federated scenario the work is motivated by, and it should not be read as a federated result.

### 4.3 Qualitative Explanations

Figures 6 and 7 show Grad-CAM and SmoothGrad maps of the centralized model and of FedProx at $\alpha = 0.1$ for six classes. The image for each class is the first one in the fixed sample, chosen before any attribution was computed, so the examples are not selected for appearance. Grad-CAM maps of both models are coarse, a consequence of the 7 × 7 feature resolution, and usually cover a large part of the leaf rather than individual lesions. For several images the two models emphasize different regions: for grape black rot the centralized map concentrates below the leaf center while the federated map spreads over the central leaf, and for corn common rust the centralized map peaks at the left edge of the image. SmoothGrad maps are finer and highlight some individual spots, for example on the apple scab leaf, but they are noisier for the federated model and also respond to leaf texture and edges.

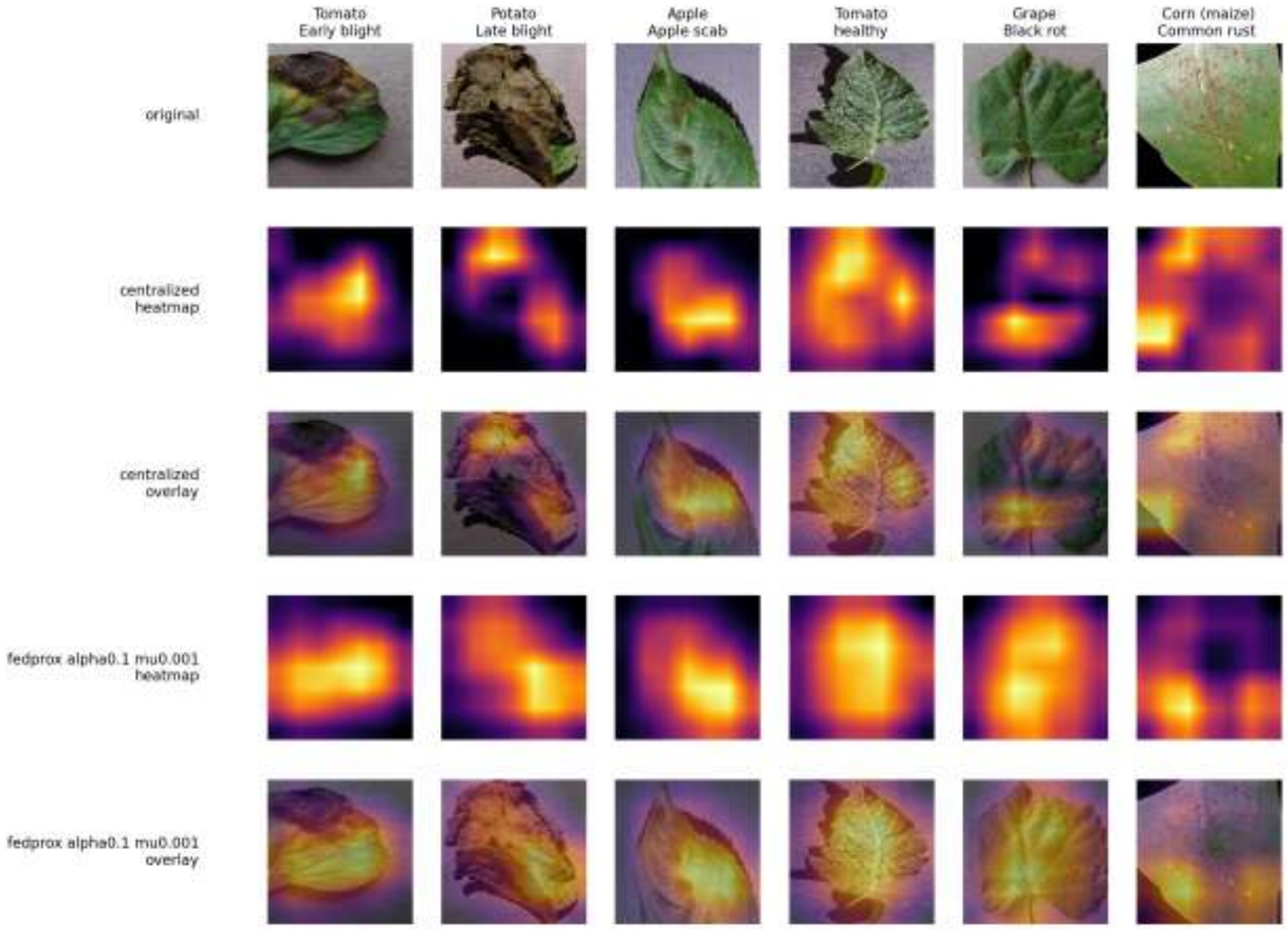

**Figure 6.** Grad-CAM for six classes. Row 1: input images. Rows 2 and 3: heatmap and overlay for the centralized model. Rows 4 and 5: heatmap and overlay for the FedProx global model ($\alpha = 0.1$, $\mu = 0.001$, seed 0). Brighter colours indicate higher attribution.

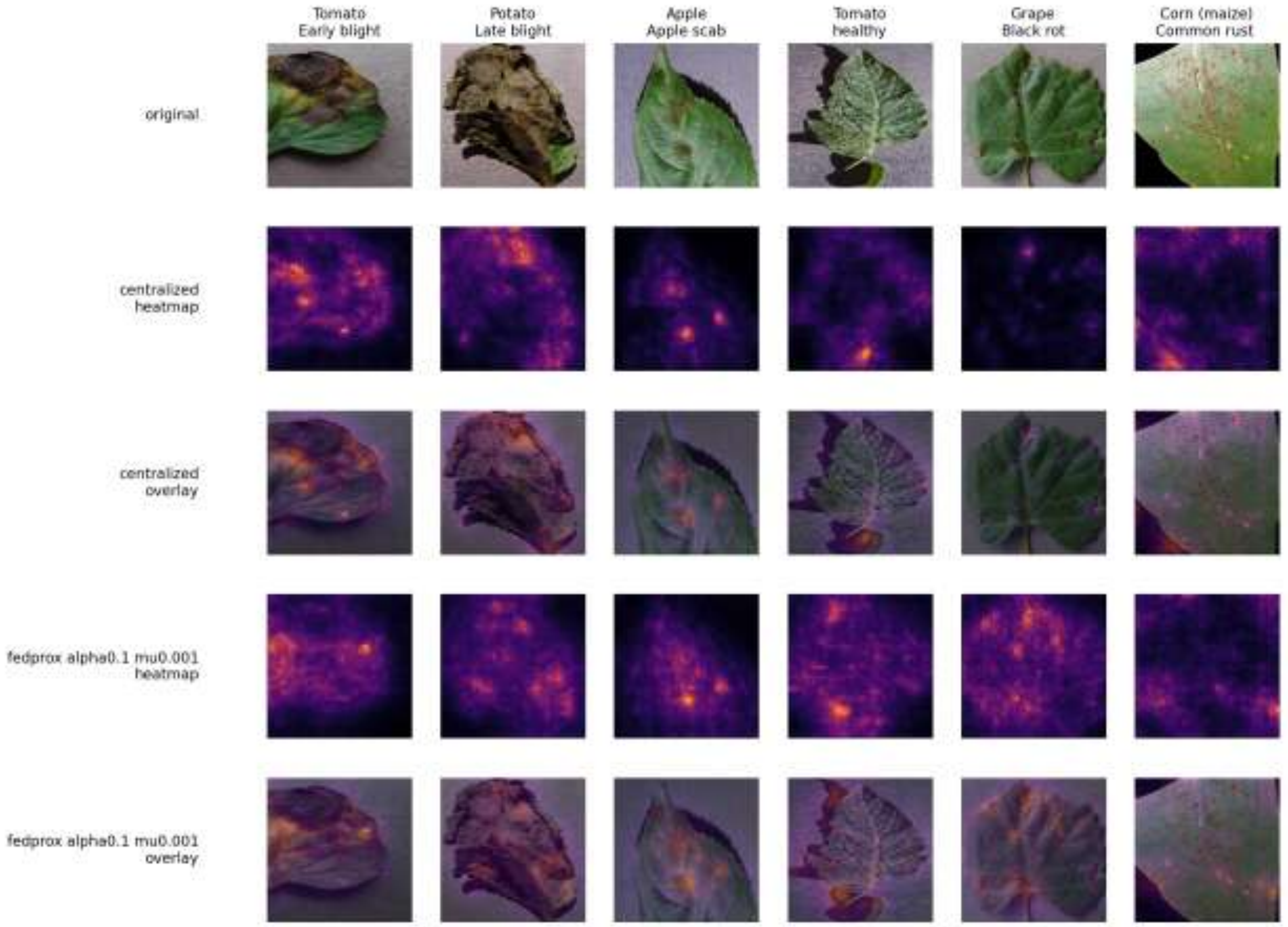

**Figure 7.** SmoothGrad for the same images and models as Figure 6. Row 1: input images. Rows 2 and 3: centralized model. Rows 4 and 5: FedProx global model ($\alpha$ = 0.1, $\mu$ = 0.001, seed 0).

### 4.4 Quantitative Evaluation of Explanations

Table 4 reports the explanation metrics for all five models and both methods.

*Table 4. Explanation metrics on the 380-image sample (lesion metrics on the 250 images with a non-empty lesion pseudo-mask). Del.: deletion AUC (lower is better). Ins.: insertion AUC (higher is better). Energy: share of attribution inside the mask. Point.: pointing game hit rate. ρ and IoU: agreement with the centralized model's map. All federated models are seed 0. Chance levels for the localization metrics are the mean mask area fractions.*

| Method / model | Del. | Ins. | Leaf energy | Leaf point. | Lesion energy | Lesion point. | Spearman ρ | Top-20% IoU |
|---|---|---|---|---|---|---|---|---|
| Grad-CAM | | | | | | | | |
| Centralized | 0.393 | 0.783 | 0.777 | 0.958 | 0.217 | 0.320 | n/a | n/a |
| FedAvg, $\alpha$ = 0.1 | 0.332 | 0.743 | 0.760 | 0.979 | 0.202 | 0.284 | 0.739 | 0.440 |
| FedProx, $\alpha$ = 0.1 | 0.319 | 0.753 | 0.765 | 0.982 | 0.204 | 0.312 | 0.735 | 0.439 |
| FedAvg, $\alpha$ = 0.5 | 0.362 | 0.778 | 0.761 | 0.987 | 0.203 | 0.300 | 0.751 | 0.451 |
| FedProx, $\alpha$ = 0.5 | 0.366 | 0.768 | 0.763 | 0.976 | 0.203 | 0.280 | 0.760 | 0.459 |
| SmoothGrad | | | | | | | | |
| Centralized | 0.294 | 0.759 | 0.700 | 0.816 | 0.200 | 0.616 | n/a | n/a |
| FedAvg, $\alpha$ = 0.1 | 0.270 | 0.714 | 0.736 | 0.921 | 0.204 | 0.608 | 0.622 | 0.305 |

| Method / model | Del. | Ins. | Leaf energy | Leaf point. | Lesion energy | Lesion point. | Spearman ϱ | Top-20% IoU |
|---|---|---|---|---|---|---|---|---|
| FedProx, $\alpha = 0.1$ | 0.291 | 0.727 | 0.738 | 0.929 | 0.203 | 0.636 | 0.611 | 0.300 |
| FedAvg, $\alpha = 0.5$ | 0.282 | 0.790 | 0.729 | 0.900 | 0.208 | 0.640 | 0.659 | 0.332 |
| FedProx, $\alpha = 0.5$ | 0.287 | 0.783 | 0.724 | 0.908 | 0.208 | 0.640 | 0.669 | 0.339 |
| Chance level | | | 0.601 | 0.601 | 0.156 | 0.156 | | |

Half-widths of the 95% bootstrap intervals over images: 0.01 to 0.03 for the AUCs and energy ratios, 0.01 to 0.04 for leaf pointing, 0.05 to 0.06 for lesion pointing and 0.01 to 0.02 for agreement.

**Leaf localization.** Every model and method placed more attribution on the leaf than chance. The Grad-CAM energy ratio was 0.760 to 0.777 and its peak fell on the leaf for 95.8 to 98.7 percent of images, against a chance level of 0.601. SmoothGrad was less concentrated (0.700 to 0.738). The federated SmoothGrad maps were more leaf-focused than the centralized ones (pointing 90.0 to 92.9 percent against 81.6 percent), so federated training did not reduce leaf-level localization.

**Lesion localization.** Attribution on lesion tissue was only modestly above chance. The lesion energy ratio was 0.200 to 0.217 for every model and method, against a lesion area fraction of 0.156. Grad-CAM's peak fell inside the lesion pseudo-mask for 28.0 to 32.0 percent of images and SmoothGrad's for 60.8 to 64.0 percent, against a chance rate of about 15.6 percent. The higher SmoothGrad rate should be read together with the sanity check below, since lesions are regions of strong colour and texture contrast to which an image-driven method would also respond.

**Faithfulness.** Deletion and insertion favour different conclusions. For the centralized model, SmoothGrad has the lower (better) deletion AUC (0.294 against 0.393) while Grad-CAM has the higher (better) insertion AUC (0.783 against 0.759). The federated models have lower Grad-CAM deletion AUCs than the centralized model (0.319 to 0.366 against 0.393). Deletion and insertion scores depend on how sensitive each model is to perturbed inputs as well as on its attribution maps, so differences between models do not isolate explanation quality, and no random-attribution baseline was computed. These metrics are therefore used here only to compare methods within a model. Per-image distributions of deletion AUC, insertion AUC and leaf energy ratio are shown in Supplementary Figures S2 to S4.

**Agreement with the centralized model.** Federated Grad-CAM maps correlated with the centralized maps at $\varrho = 0.735$ to 0.760, with 44 to 46 percent overlap of the top 20 percent of pixels; SmoothGrad agreement was lower ($\varrho = 0.611$ to 0.669, overlap 30 to 34 percent). For both methods agreement was lower at $\alpha = 0.1$ than at $\alpha = 0.5$ (Figure 8), so heterogeneity moved the federated explanations further from the centralized ones. Explanations were therefore partly, but not fully, preserved through aggregation.

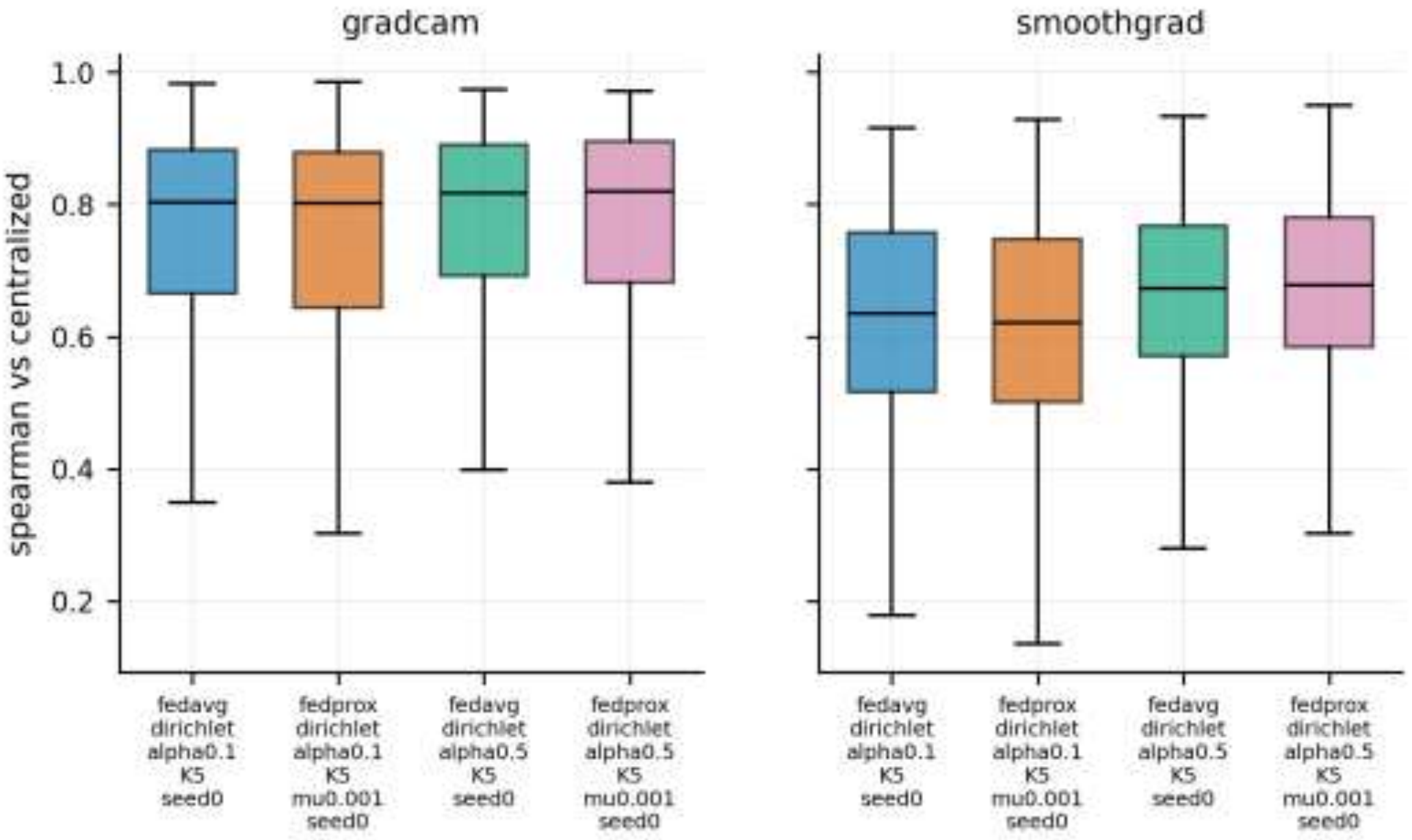

**Figure 8.** Per-image Spearman correlation between each federated model's attribution map and the centralized model's map on the 380-image sample. Boxes show the interquartile range and median; whiskers extend to 1.5 times the interquartile range, and outliers are hidden.

**FedProx against FedAvg.** Differences between the two strategies were small and inconsistent in direction. The largest median paired difference on any metric was 0.013. Some differences were significant because the 380 paired images give high power, but they point in both directions: FedProx agreed more with the centralized SmoothGrad maps at $\alpha = 0.5$ (ρ 0.669 against 0.659, $p < 0.0001$) and less at $\alpha = 0.1$ (0.611 against 0.622, $p = 0.0002$), and its Grad-CAM agreement was higher at $\alpha = 0.5$ (0.760 against 0.751, $p = 0.010$) but not different at $\alpha = 0.1$ ($p = 0.30$). Supplementary Table S1 lists all 32 tests. Across them, without correction for multiple comparisons, no consistent advantage emerges, and the hypothesis that proximal regularization preserves explanation quality better than FedAvg is not supported in this setting.

**Sanity check.** Table 5 shows how attribution maps of the centralized model change as its weights are randomized. Grad-CAM maps changed substantially once the classifier was randomized (ρ = 0.39), and the correlation fell to between 0.15 and 0.18 once the middle and early blocks were also randomized. SmoothGrad maps remained highly correlated with the original after the classifier and the last feature block were randomized (ρ = 0.75 and 0.81), and fell to 0.20 to 0.32 only after earlier blocks were also randomized. SmoothGrad therefore partly fails this check: much of its map is determined by low-level image structure, independent of what the classifier learned. Its localization scores, including its higher lesion pointing rate, should be interpreted with this in mind.

*Table 5. Model randomization check on the centralized model (40 images from the four apple classes). Mean Spearman correlation between the map of the trained model and the map after cumulative randomization of the layers listed and all layers above them. Values near zero indicate that the method depends on the learned weights.*

| Layers randomized (cumulative) | Grad-CAM ρ | SmoothGrad ρ |
|---|---|---|
| Classifier | 0.392 | 0.746 |
| + features[18] | 0.365 | 0.808 |
| + features[14:18] | 0.291 | 0.321 |
| + features[10:14] | 0.145 | 0.234 |
| + features[6:10] | 0.178 | 0.202 |

| Layers randomized (cumulative) | Grad-CAM ϱ | SmoothGrad ϱ |
|---|---|---|
| + features[0:6] (whole network) | 0.177 | 0.256 |

### 4.5 Failure Cases

Figure 9 shows the six sampled images on which the FedProx model at $\alpha = 0.1$ placed the least attribution on the leaf (mean leaf energy ratio of the two methods between 0.29 and 0.44). Five of the six show a leaf that occupies a small or narrow part of the frame: a small tomato leaf with Septoria leaf spot, a shrivelled tomato leaf with late blight, and three narrow healthy peach leaves. The sixth is a tomato late blight leaf with extensive dark necrotic tissue. Grad-CAM's coarse resolution spreads attribution into the background around small and narrow leaves, and SmoothGrad also assigns attribution to background texture. These explanations are least reliable when the leaf fills little of the image.

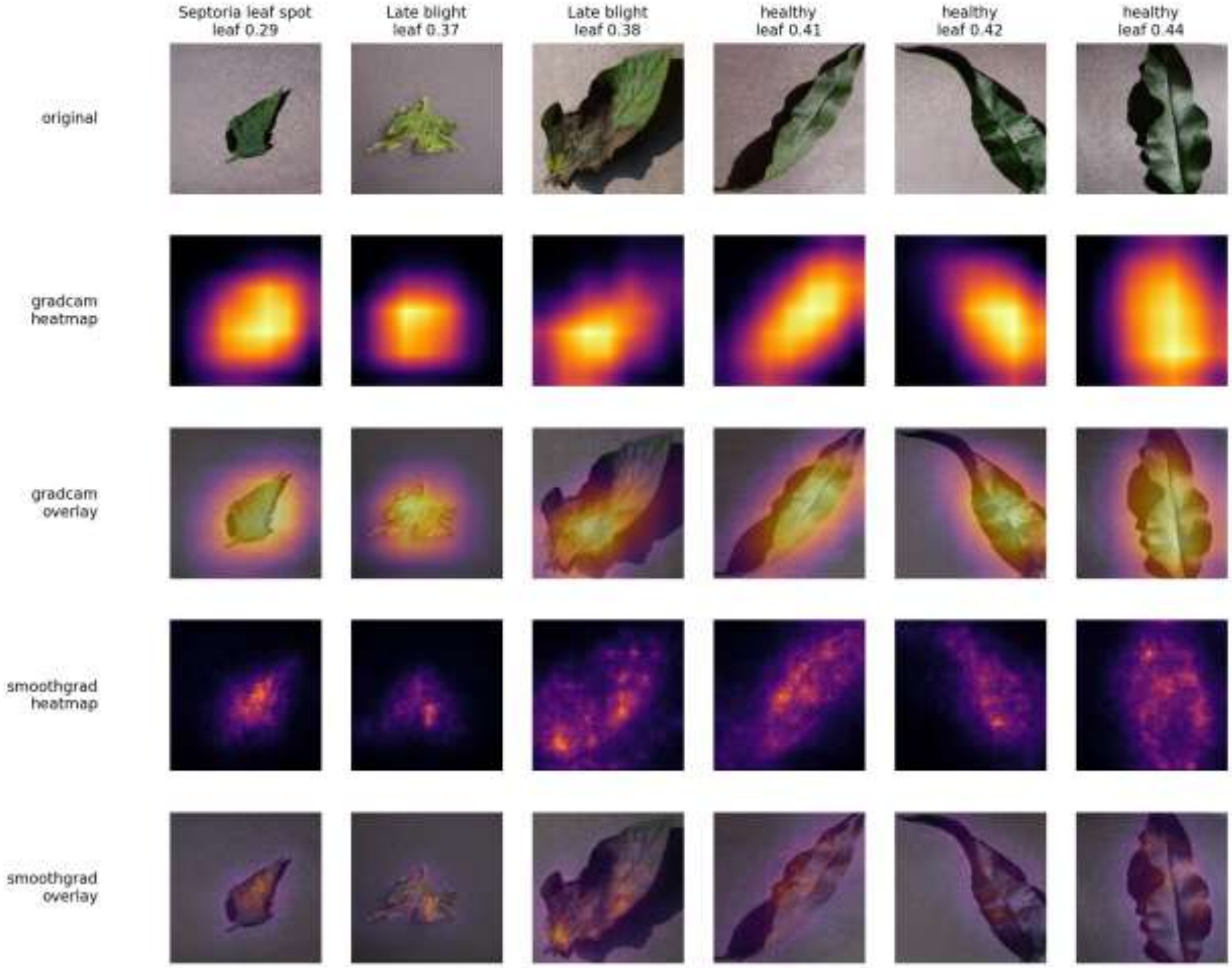

**Figure 9.** The six sampled images with the lowest leaf energy ratio for the FedProx global model ($\alpha = 0.1$, $\mu = 0.001$, seed 0). The value above each column is the leaf energy ratio averaged over the two methods. Rows show the input, Grad-CAM heatmap and overlay, and SmoothGrad heatmap and overlay.

## 5. Discussion

**Accuracy.** In this setting, federated training cost 0.57 percentage points of accuracy with identically distributed clients and up to 1.60 points under strong label skew ($\alpha = 0.1$). The penalty grows with heterogeneity, but at moderate skew ($\alpha = 0.5$ and $\alpha = 1$) it was not distinguishable from the IID case on the paired test predictions. FedProx gave no measurable benefit: the coefficient selected on validation was the smallest one tried, and a larger coefficient significantly reduced accuracy. FedProx may still help in regimes not studied here, such as several local epochs per round, partial client participation or many more clients, all of which increase client drift.

**Explanations.** Post-aggregation Grad-CAM maps remained concentrated on the leaf and correlated well with those of the centralized model, but agreement decreased with heterogeneity, and attribution on lesion tissue was only modestly above chance. These maps support the statement that the federated models attend to the leaf rather than to the background. They do not support stronger statements that the models localize individual lesions or that their explanations are equivalent to those of a centrally trained model. SmoothGrad produced finer and more lesion-like maps, but the randomization check shows that much of that structure comes from the image rather than from the classifier, which is a reason not to present SmoothGrad maps to practitioners as evidence of what the model learned. More broadly, the results show why visual inspection of a few examples is not sufficient: the qualitative grids in Figures 6 and 7 look plausible, while the quantitative evaluation reveals the limits of both methods.

**Limitations.** Several limitations apply. (1) PlantVillage images were captured under controlled conditions with uniform backgrounds, and 2.89 percent of test images have a near-duplicate in the training split, so absolute accuracies are optimistic for field use. (2) Heterogeneity was simulated as label skew among five clients with full participation; real farms also differ in imaging devices, lighting and backgrounds (feature shift), which was not modelled. (3) The pipeline provides no formal privacy guarantee: no secure aggregation or differential privacy was used, and model updates can leak information about training images [60]. (4) The lesion pseudo-masks are a colour heuristic (Figure 3). They miss lesions whose hue lies within the healthy range (6 of the 10 sampled tomato mosaic virus images had empty masks) and can mark whole leaves whose colour changes overall (for orange Huanglongbing, the median pseudo-mask covers 40 percent of the image), so lesion metrics are approximate. (5) Explanations were evaluated for one seed per configuration, deletion and insertion were not compared with a random-attribution baseline, and the explanation tests were not corrected for multiple comparisons.

## 6. Conclusion and Future Work

This paper presented FedXCrop, a federated pipeline for crop disease classification in which the explanations of aggregated models are evaluated quantitatively. On PlantVillage, a centrally trained MobileNetV2 reached 99.38 ± 0.05 percent test accuracy. Federated models trained from ImageNet weights on the same data reached 98.81 ± 0.09 percent with identically distributed clients and 97.78 ± 0.48 percent under strong label skew, and FedProx did not improve on FedAvg at any level of heterogeneity. Grad-CAM explanations of the aggregated models stayed on the leaf and correlated with those of the centralized model, with agreement decreasing as heterogeneity increased. Lesion-level localization was weak, and SmoothGrad partly failed a model randomization check. Federated training thus retains most of the accuracy of centralized training and much of its Grad-CAM behaviour, but explanations from either training regime should be validated quantitatively before they are used to justify treatment decisions.

Future work will evaluate the pipeline on field-captured imagery such as PlantDoc [62], with feature shift between clients as well as label skew; study larger client populations, partial participation and more local epochs, where proximal regularization may matter; add secure aggregation and differential privacy and measure their effect on accuracy and explanations; and replace colour-based pseudo-masks with expert-annotated lesion masks.

## Declarations

### Data Availability

The PlantVillage dataset [2,61] is publicly available. The copy used in this study, containing the colour, segmented and grayscale variants, was obtained from Kaggle: https://www.kaggle.com/datasets/amgadalameri/plantvillage-dataset-zip

**Code Availability**

The code, configuration files, data split, client partitions and all result files from which the tables and figures were produced are available at https://github.com/Kriti2709/fedxcrp

**Competing Interests**

The authors declare that they have no competing interests.

**Funding**

The authors did not receive funding for this study.

**Authors' Contributions**

All authors contributed to the conception and design of the study. Kriti Saini and Manikandan P performed the analysis, and Manikandan P and Omana J interpreted the findings. All authors contributed to the preparation of the manuscript and read and approved the final version.

**Acknowledgements**

The authors acknowledge the support of the Vellore Institute of Technology, Chennai.

**Consent to Publish**

Not applicable.

**Ethics and Consent to Participate**

Not applicable.

**Supplementary Material**

The figures and table below support statements in the main text. All are produced by the released code from the stored result files.

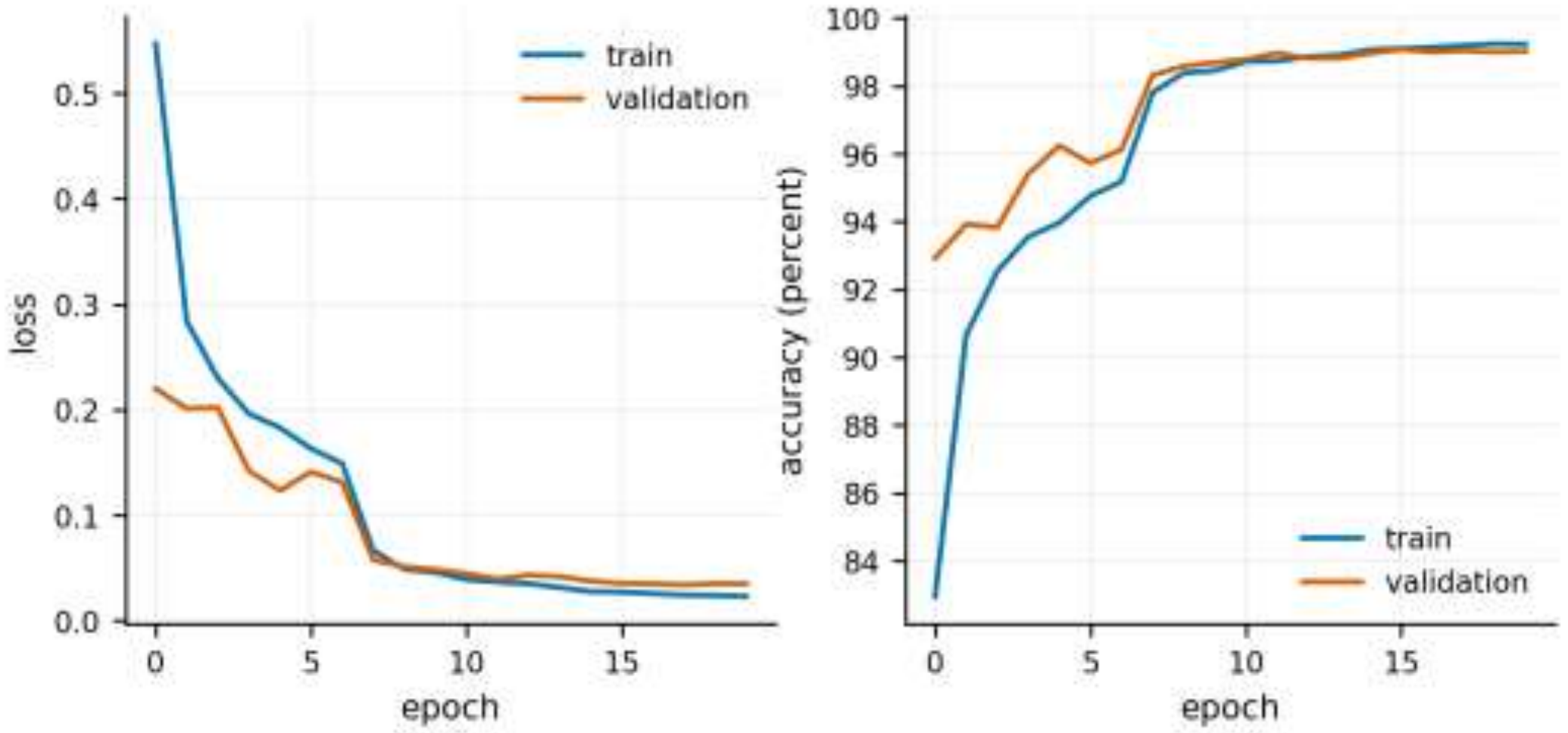

**Figure S1.** Training and validation loss (left) and accuracy (right) of the centralized model, seed 0. Epochs are counted from 0. The step in both curves after epoch 6 is the learning rate decay. Validation accuracy exceeds training accuracy early in training because augmentation is applied only to the training split. The curves for seeds 1 and 2 are stored with the results.

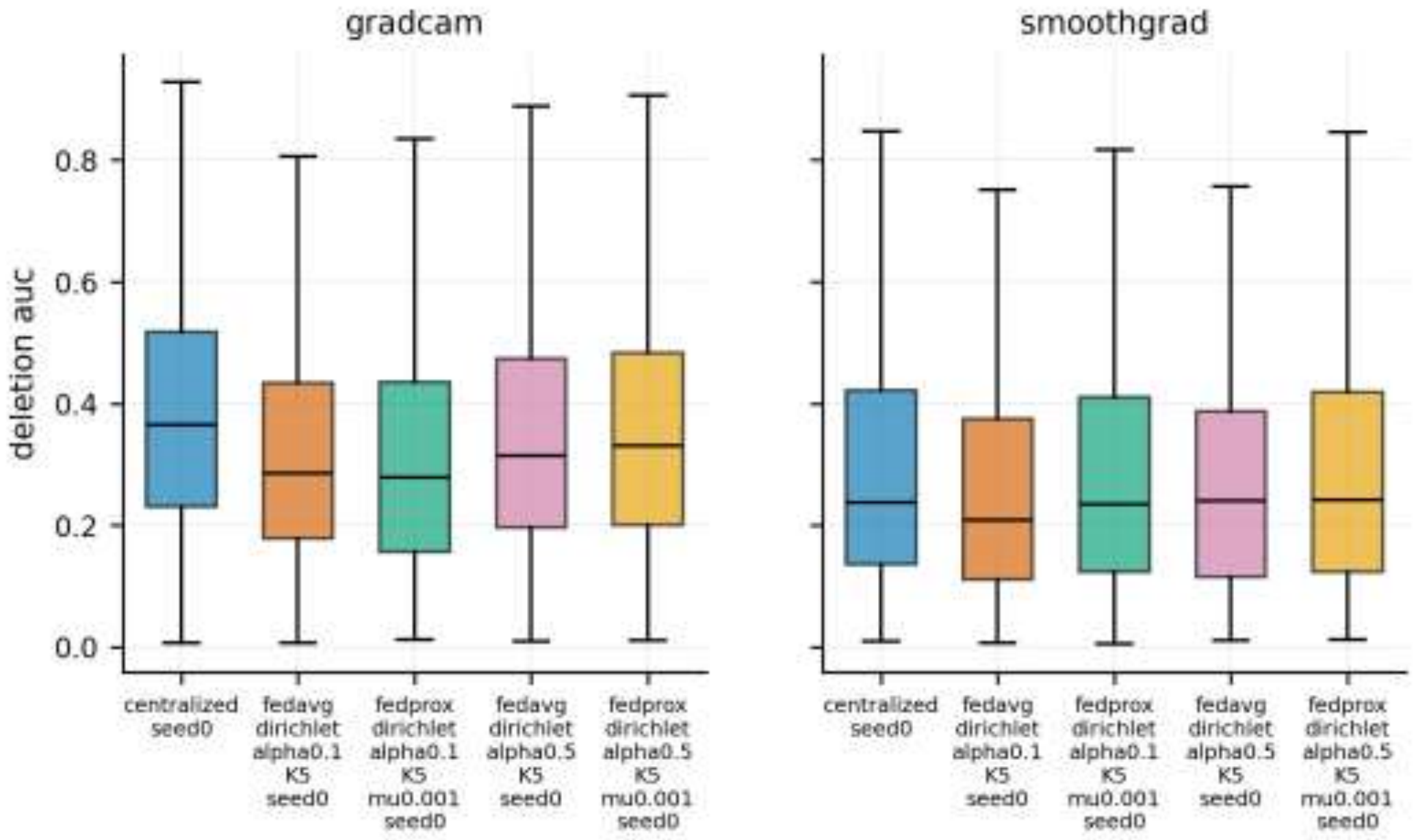

**Figure S2.** Per-image deletion AUC on the 380-image sample for each model and method (lower is better). Boxes show the interquartile range and median; whiskers extend to 1.5 times the interquartile range, and outliers are hidden.

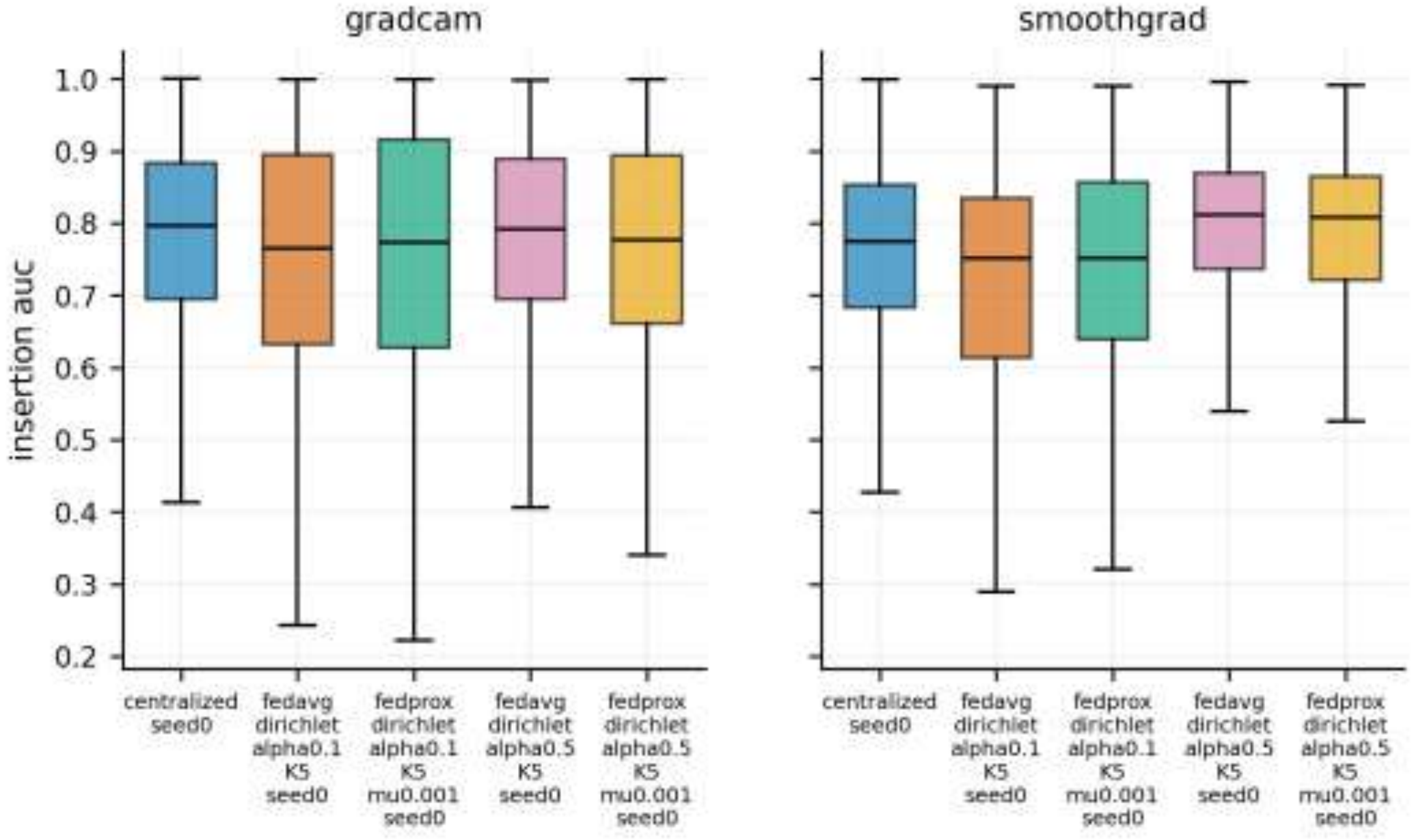

**Figure S3.** Per-image insertion AUC on the 380-image sample (higher is better). Layout as in Figure S2.

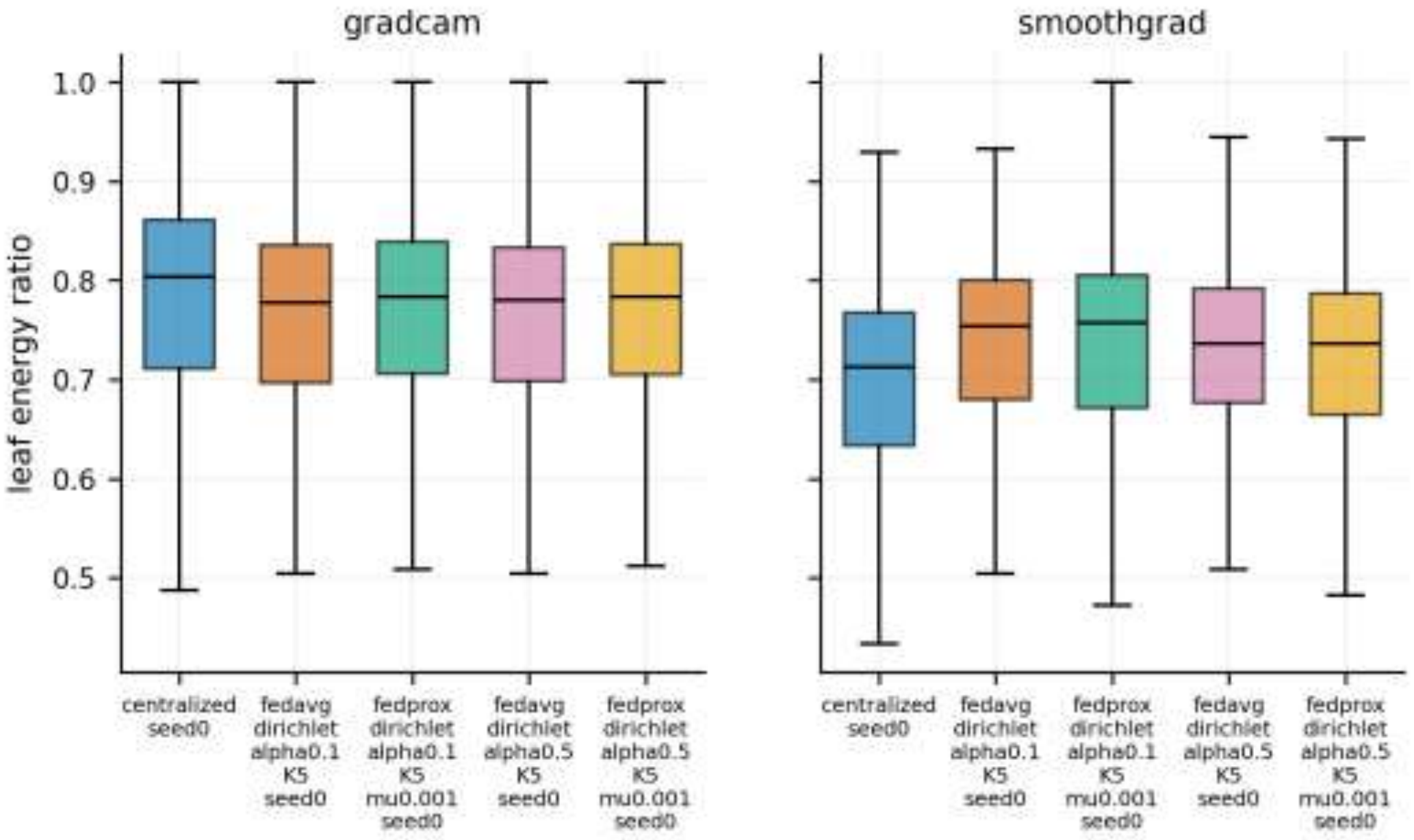

**Figure S4.** Per-image leaf energy ratio on the 380-image sample. The chance level, the mean leaf area fraction, is 0.601. Layout as in Figure S2.

*Table S1. Paired Wilcoxon signed-rank tests of FedProx (μ = 0.001) against FedAvg on identical images, seed 0 models. Means are over images; the median difference is FedProx minus FedAvg. Lower deletion AUC is better; for every other metric higher is better. No correction for multiple comparisons is applied.*

| Method | α | Metric | FedProx mean | FedAvg mean | Median difference | Pairs | p |
|---|---|---|---|---|---|---|---|
| Grad-CAM | 0.1 | Deletion AUC | 0.319 | 0.332 | -0.0051 | 380 | 0.0027 |
| Grad-CAM | 0.1 | Insertion AUC | 0.753 | 0.743 | +0.0130 | 380 | 0.0115 |
| Grad-CAM | 0.1 | Leaf energy | 0.765 | 0.760 | +0.0044 | 380 | < 0.0001 |
| Grad-CAM | 0.1 | Leaf pointing | 0.982 | 0.979 | +0.0000 | 380 | 0.7389 |
| Grad-CAM | 0.1 | Lesion energy | 0.204 | 0.202 | +0.0004 | 250 | 0.0571 |
| Grad-CAM | 0.1 | Lesion pointing | 0.312 | 0.284 | +0.0000 | 250 | 0.2230 |
| Grad-CAM | 0.1 | Spearman ρ | 0.735 | 0.739 | -0.0017 | 380 | 0.2954 |
| Grad-CAM | 0.1 | Top-20% IoU | 0.439 | 0.440 | +0.0012 | 380 | 0.8480 |
| Grad-CAM | 0.5 | Deletion AUC | 0.366 | 0.362 | +0.0011 | 380 | 0.2532 |
| Grad-CAM | 0.5 | Insertion AUC | 0.768 | 0.778 | -0.0054 | 380 | 0.0253 |
| Grad-CAM | 0.5 | Leaf energy | 0.763 | 0.761 | +0.0022 | 380 | 0.0020 |
| Grad-CAM | 0.5 | Leaf pointing | 0.976 | 0.987 | +0.0000 | 380 | 0.1025 |
| Grad-CAM | 0.5 | Lesion energy | 0.203 | 0.203 | +0.0001 | 250 | 0.8616 |
| Grad-CAM | 0.5 | Lesion pointing | 0.280 | 0.300 | +0.0000 | 250 | 0.4838 |
| Grad-CAM | 0.5 | Spearman ρ | 0.760 | 0.751 | +0.0043 | 380 | 0.0100 |
| Grad-CAM | 0.5 | Top-20% IoU | 0.459 | 0.451 | +0.0048 | 380 | 0.0561 |
| SmoothGrad | 0.1 | Deletion AUC | 0.291 | 0.270 | +0.0130 | 380 | < 0.0001 |
| SmoothGrad | 0.1 | Insertion AUC | 0.727 | 0.714 | +0.0065 | 380 | 0.0071 |

| **Method** | **$\alpha$** | **Metric** | **FedProx mean** | **FedAvg mean** | **Median difference** | **Pairs** | **p** |
|---|---|---|---|---|---|---|---|
| SmoothGrad | 0.1 | Leaf energy | 0.738 | 0.736 | +0.0016 | 380 | 0.0674 |
| SmoothGrad | 0.1 | Leaf pointing | 0.929 | 0.921 | +0.0000 | 380 | 0.6473 |
| SmoothGrad | 0.1 | Lesion energy | 0.203 | 0.204 | -0.0007 | 250 | 0.0490 |
| SmoothGrad | 0.1 | Lesion pointing | 0.636 | 0.608 | +0.0000 | 250 | 0.3363 |
| SmoothGrad | 0.1 | Spearman ϱ | 0.611 | 0.622 | -0.0056 | 380 | 0.0002 |
| SmoothGrad | 0.1 | Top-20% IoU | 0.300 | 0.305 | -0.0048 | 380 | 0.0073 |
| SmoothGrad | 0.5 | Deletion AUC | 0.287 | 0.282 | +0.0075 | 380 | 0.0443 |
| SmoothGrad | 0.5 | Insertion AUC | 0.783 | 0.790 | -0.0058 | 380 | 0.0308 |
| SmoothGrad | 0.5 | Leaf energy | 0.724 | 0.729 | -0.0029 | 380 | < 0.0001 |
| SmoothGrad | 0.5 | Leaf pointing | 0.908 | 0.900 | +0.0000 | 380 | 0.6394 |
| SmoothGrad | 0.5 | Lesion energy | 0.208 | 0.208 | +0.0001 | 250 | 0.3237 |
| SmoothGrad | 0.5 | Lesion pointing | 0.640 | 0.640 | +0.0000 | 250 | 1.0000 |
| SmoothGrad | 0.5 | Spearman ϱ | 0.669 | 0.659 | +0.0087 | 380 | < 0.0001 |
| SmoothGrad | 0.5 | Top-20% IoU | 0.339 | 0.332 | +0.0053 | 380 | 0.0012 |